\documentclass{ijac}
\usepackage{multicol}
\usepackage{caption}
\usepackage{subcaption}
\usepackage{amsmath}
\usepackage{amssymb}
\usepackage{amsthm}
\usepackage{amsfonts}
\usepackage{graphicx}
\usepackage{url}
\usepackage{ccaption}
\usepackage{booktabs}

\usepackage{times}
\usepackage{helvet}
\usepackage{courier}
\usepackage{amssymb,amsfonts,amsmath,dsfont,amsthm}
\usepackage[utf8]{inputenc} 
\usepackage[T1]{fontenc}    
\usepackage{url}            
\usepackage{booktabs}       
\usepackage{nicefrac}       
\usepackage{microtype}      
\usepackage[svgnames]{xcolor}
\usepackage{xcolor,colortbl}
\providecommand{\scal}[2]{\left\langle{#1},{#2}\right\rangle}

\newcommand{\BR}{\mathbb{R}}
\newcommand{\ra}{\rightarrow}
\newcommand{\hh}{\mathcal{H}}
\newcommand{\F}{\mathcal{F}}
\newcommand{\cX}{\mathcal{X}}

\DeclareMathOperator{\R}{\mathbb{R}}
\providecommand{\nor}[1]{\left\lVert {#1} \right\rVert}
\providecommand{\scal}[2]{\left\langle{#1},{#2}\right\rangle}
\newcommand{\be}{\begin{equation}}
\newcommand{\ee}{\end{equation}}
\newcommand{\bt}{\begin{theorem}}
\newcommand{\et}{\end{theorem}}
\newcommand{\bd}{\begin{definition}}
\newcommand{\ed}{\end{definition}}
\newcommand{\br}{\begin{remark}}
\newcommand{\er}{\end{remark}}

\newtheorem{theorem}{Theorem}
\newtheorem{definition}{Definition}
\newtheorem{conjecture}{Conjecture}
\newtheorem{lemma}{Lemma}
\newtheorem{proposition}{Proposition}
\newtheorem{remark}{Remark}

\newtheorem{corollary}{Corollary}
\newtheorem{prob}{Problem}

\newcommand{\suchthat}{\, \mid \,} 

\firstheadname{} 

\firstfootname{} 
\headevenname{} 
\headoddname{}

\renewcommand{\baselinestretch}{1}
\newcommand\blfootnote[1]{%
  \begingroup
  \renewcommand\thefootnote{}\footnote{#1}%
  \addtocounter{footnote}{-1}%
  \endgroup
}

\newcommand*{\titleAT}{\begingroup
  \newlength{\drop}
  \drop=0.05\textheight
  \begin{center}
  \includegraphics[scale=0.4]{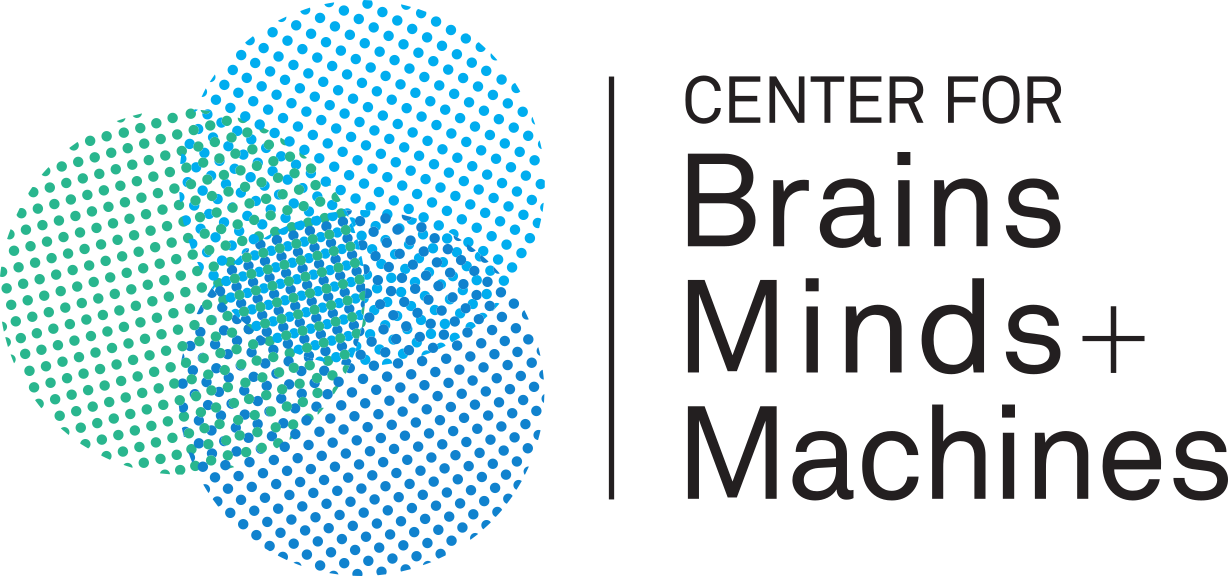}
  \end{center}
  \vspace{2pt}\vspace{-\baselineskip}

  \textbf{\large{CBMM Memo No. \memonumber}}   \hfill    \textbf{\large{\memodate}}

  \vspace{\drop}
  \begin{center}
    \textbf{\huge{\memotitle}}\\
    \vspace{0.4\drop}
    \textbf{\Large{by}}\\
    \vspace{0.4\drop}
    \large{\memoauthors}
  \end{center}
  \vspace{\drop}
  \textbf{\large{\noindent Abstract}:} {\memoabstract}

\blfootnote{This memo reports those parts of CBMM Memo 067 that are focused on
  optimization. The main reason is to straighten up the titles of the
  theory trilogy.}

\vspace{\fill}
  \rule{\textwidth}{0.4pt}\par

  \begin{minipage}{.15\linewidth}
    \includegraphics[scale=0.1]{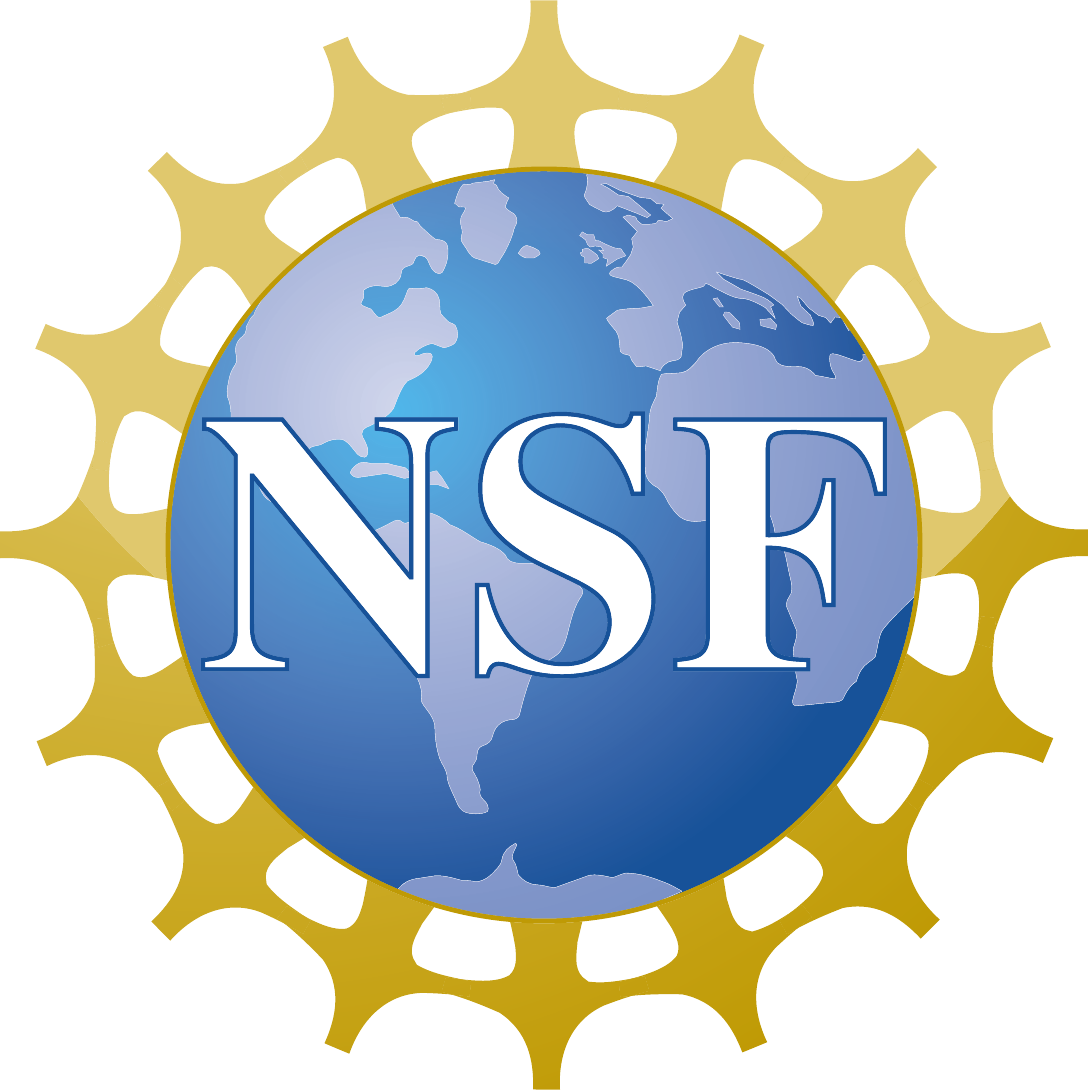}
  \end{minipage}
  \begin{minipage}{.84\linewidth}
    \textbf{\large{This work was supported by the Center for Brains,
        Minds and Machines (CBMM), funded by NSF STC award  CCF - 123
1216.
    }}
  \end{minipage}
  \endgroup}

\def\RR{{\mathbb R}}

\def\PPI{{{\rm I}\kern-1pt\Pi}}

\def\b #1;{{\bf #1}}

\def\F{{\cal F}}

\def\P{{\cal P}}

\def\Z{{\cal Z}}

\def\E{{\mathbb E}}

\def\Y{{\bf Y}}

\def\be{\begin{equation}}
\def\ee{\end{equation}}
\def\bea{\begin{eqnarray}}
\def\eea{\end{eqnarray}}

\def\donchitre#1#2{\vskip 6.5cm\noindent
\parbox[t]{1in}{\special{eps:#1.eps x=6.5cm y=5.5cm}}
\hbox to 7cm{}\parbox[t]{0.0cm}{\special{eps:#2.eps x=6.5cm y=5.5cm}}}

\begin{document}

\setcounter{page}{1}
\pagenumbering{arabic}
\onecolumn
\def\memonumber{072}
\def\memodate{\today}
\def\memotitle {Theory of Deep Learning IIb: Optimization
  Properties of SGD}

\def\memoauthors{Chiyuan Zhang$^{1}$ \quad Qianli Liao$^{1}$ \quad
  Alexander Rakhlin$^2$ 
\quad Brando
Miranda$^1$
\quad Noah Golowich$^1$ \quad Tomaso Poggio$^{1}$ \\
\vskip 0.3in
$^1$Center for Brains, Minds, and Machines, McGovern Institute for Brain Research, \\
   Massachusetts Institute of Technology, Cambridge, MA, 02139. \\
   $^2$ University of Pennsylvania \\
}

\normalsize \def\memoabstract {In Theory IIb we characterize with a
  mix of theory and experiments the optimization  of deep
  convolutional networks by Stochastic
  Gradient Descent. The main new result in this paper is theoretical
  and experimental evidence for the following conjecture about SGD:
{\it SGD concentrates in probability - like the classical Langevin
  equation -- on large volume, ``flat'' minima, selecting flat
  minimizers which are with very high probability also {\it global minimizers}}.
}

\titleAT
\newpage

\title
{Theory of Deep Learning IIb: Optimization
  Properties of SGD}

\author{
Chiyuan Zhang$^1$ \qquad Qianli Liao$^{1}$  \qquad
  Noah Golowich$^{1}$  \qquad Sasha Rakhlin$^2$ \qquad Karthik Sridharan \qquad Tomaso Poggio$^{1}$\\ }

\address{
$^1$Center for Brains, Minds, and Machines, McGovern Institute for Brain Research,
   Massachusetts Institute of Technology, Cambridge, MA, 02139.
\\
$^2$Department of Statistics, Wharton.

}



\pagestyle{ijacheadings}

\section{Introduction}

In the last few years, deep learning has been tremendously successful
in many important applications of machine learning. However, our
understanding of deep learning is still far from complete. A
satisfactory characterization of deep learning should cover the
following parts: 1) representation power --- what types of functions
can neural networks (DNNs) represent well and what are the
advantages and disadvatages  over using shallow models? 2) optimization of the empirical
loss --- can we characterize the convergence of stochastic gradient
descent (SGD) on the non-convex empirical loss encountered in deep
learning? 3) why do the deep learning models,
despite being highly over-parameterized, still predict well?

The first two questions are addressed in Theory I \cite{Theory_I} and
Theory II \cite{Theory_II} respectively. In this paper, we add new
results to the second question. In the rest of the paper, we try to address this set of issues at the
level of formal rigor of a physicist (not a mathematician: this will
come later). 

\subsection{Related work}

Recent work by Keskar et al.\cite{keskar_large-batch_2016} is 
relevant. The authors estimate the loss in a neighborhood of the
weights to argue that small batch size in SGD (i.e., larger gradient
estimation noise, see later) generalizes better than large
mini-batches and also results in significantly flatter minima. In
particular, they note that the stochastic gradient descent method used
to train deep nets, operate in a small-batch regime wherein a fraction
of the training data, usually between $32$ and $512$ data points, is
sampled to compute an approximation to the gradient. They discuss the
common observation that when using a larger batch there is a
significant degradation in the quality of the model, as measured by
its ability to generalize. We provide theoretical arguments for the
cause for this generalization drop in the large-batch regime,
supporting the numerical evidence of Keskar et al.. The latter shows
that large-batch methods tend to converge to sharp minimizers of the
training and testing functions --- and that sharp minima lead to poor
generalization. In contrast, small-batch methods consistently converge
to minimizers that generalize better, and our theoretical arguments
support a commonly held view that this is due to the inherent noise in
the gradient estimation. Notice however that our explanation is
related to optimization rather than generalization.

On the other hand, as shown in \cite{dinh2017sharp}, sharp minimizers
do not necessarily lead to bad generalization performance. Due to the
parameterization redundancy in deep neural networks, given any (flat)
minimizers, one can artificially transform the weights to land in a
sharp but equivalent minimizer because the function defined by the
weights are the same. Notice that the argument in \cite{dinh2017sharp}
does not conflict with the argument that flat minimizers generalize
well. Moreover, isotropic flat minima in the loss wrt {\it all
  weights}, if they exist, cannot be transformed in sharp minima.

\subsection{Landscape of the Empirical Risk}
\label{Landscape}

In Theory II \cite{Theory_II} we have described some features of the
landscape of the empirical risk, for the case of deep networks of the
compositional type (with weight sharing, though the proofs do not need
the weight sharing assumption). We assumed over-parametrization, that
is more parameters than training data points, as most of the
successful deep networks. Under these conditions, setting the
empirical error to zero yields a system of $\epsilon$-approximating
polynomial equations that have an infinite number of solutions (for
the network weights). Alternatively, one can replace the RELUs in the
network with an approximating univariate polynomial and verify
empirically that the network behavior is essentially unchanged. The
associated system of equations allows for a large number of solutions
-- when is not inconsistent -- which are {\it degenerate, that is
  flat} in several of the dimensions (in CIFAR there are about $10^6$
unknown parameters for $6\times10^4$ equations. Notice that solutions
with zero empirical error are global minimizers. No other solution
with zero-error exists with a deeper minimum or less generic
degeneracy.  Empirically we observe (see Theory II) that
zero-minimizers correspond to  flat regions.


\section{SGD: Basic Setting}

Let $Z$ be a probability space with an unknown measure $\rho$. A training set $S_n$ is a set of
i.i.d. samples $z_i,i=1,\ldots,n$ from $\rho$. Assume a hypothesis $\mathcal{H}$ is chosen in advance
of training. Here we assume $\mathcal{H}$ is a $p$-dimensional Hilbert space, and identify
elements of $\mathcal{H}$ with $p$-dimensional vectors in $\RR^p$. A loss function is a map $V:
\mathcal{H}\times Z\rightarrow \mathbb{R}_+$. Moreover, we assume the expected loss
\begin{equation}
  I(f) = \mathbb{E}_z V(f, z)
\end{equation}
exists for all $f\in\mathcal{H}$. We consider the problem of finding a
minimizer of $I(f)$ in a closed subset $K\subset \mathcal{H}$. We
denote this minimizer by $f_K$ so that
\begin{equation}
  I(f_K) = \min_{f\in K} I(f)
\end{equation}
In general, the existence and uniqueness of a minimizer is not guaranteed unless some further
assumptions are specified.

Since $\rho$ is unknown, we are not able evaluate $I(f)$. Instead, we try to minimize the empirical
loss
\begin{equation}
  I_{S_n}(f) = \hat{\mathbb{E}}_{z\sim S_n} V(f, z)
  = \frac{1}{n}\sum_{i=1}^n V(f, z_i)
\end{equation}
as a proxy. In deep learning, the most commonly used algorithm is SGD and its variants. The basic
version of SGD is defined by the following iterations:
\begin{equation}
  f_{t+1} = \Pi_K\left(f_t - \gamma_t \nabla V (f_t, z_t) \right)
  \label{eq:sgd-proj-def}
\end{equation}
where $z_t$ is a sampled from the training set $S_n$ uniformly at random, and
$\nabla V(f_t, z_t)$ is an unbiased estimator of the full gradient of the empirical loss at $f_t$:
\[
  \mathbb{E}_{z_t\sim S_n}\left[ \nabla V(f_t, z_t) \right] = \nabla \hat{I}(f_t)
\]
$\gamma_t$ is a decreasing sequence of non-negative numbers, usually called the \emph{learning
rates} or \emph{step sizes}. $\Pi_K:\mathcal{H}\rightarrow K$ is the projection map into $K$, and
when $K=\mathcal{H}$, it becomes the identity map. It is interesting that the following equation, labeled SGDL, and
studied by several authors, including \cite{GelMit91}, seem to work as
well as or better than the usual repeat SGD used to train deep
networks, as discussed in Section 5:

\begin{equation}
f_{t+1}=f_t - \gamma_ n \nabla V(f_t, z_t) + \gamma'_t W_t.
\label{SGDL}
\end{equation}

\noindent Here $W_t$ is a standard Gaussian vector in $\BR^p$ and 
$\gamma'_t$ is a
sequence going to zero.

We consider a situation in which the expected cost function $I(f)$ can
have, possibly multiple, {\it global} minima. As argued by \cite{bottou-98x} there
are two ways to prove convergence of SGD. The first method consists of
partitioning the parameter space into several attraction basins,
assume that after a few iterations the algorithm confines the
parameters in a single attraction basin, and proceed as in the convex
case. A simpler method, instead of proving
that the function $f$ converges, proves that the cost function $I(f)$
and its gradient $\E_z \nabla V(f, z))$ converge.

Existing results show that when the learning rates decrease with an
appropriate rate, and subject to relatively mild assumptions,
stochastic gradient descent converges almost surely to a global
minimum when the objective function is convex or
pseudoconvex\footnote{In convex analysis, a pseudoconvex function is a
  function that behaves like a convex function with respect to finding
  its local minima, but need not actually be convex. Informally, a
  differentiable function is pseudoconvex if it is increasing in any
  direction where it has a positive directional derivative.}, and
otherwise converges almost surely to a local minimum. This direct
optimization shortcuts the usual discussion for batch ERM about
differences between optimizing the empirical risk on $S_n$ and the
expected risk.

Often extra-assumptions are made to ensure convergence and
generalization by SGD. Here we observe that simulations on standard
databases remain essentially unchanged if the
domain of the weights is assumed to be a torus which is compact,
because the weights remain bounded in most cases.

\section{SGD with overparametrization}

We  conjecture that  SGD, while minimizing the empirical loss also maximizes the
  volume, that is ``flatness'', of the minima.

Our argument can be loosely described as follows. The zero minimizers
are unique for $n>>W$ and  become degenerate , that is flat, for $n <<
W$. Let us caution that counting effective parameters is tricky in the
case of deep net so the inequalities above should be considered just guidelines.

We consider the steps of our argument, {\it starting with 
properties of SGD that have been so far unrecognized} from the machine
learning point of view, to the best of
our knowledge.

\subsection{SGD as an approximate Langevin equation}

We consider the usual SGD update defined by the
recursion
\begin{equation}\label{sgdK1}
f_{t+1}=f_t - \gamma_ t \nabla V(f_t, z_t),
\end{equation}
where $z_t$ is fixed,  $\nabla V(f_t, z_t)$ is the gradient of the loss
with respect to $f$ at $z_t$, and $\gamma_t$ is a suitable decreasing
sequence. When $z_t\subset [n]$ is a minibatch, we overload the notation and write $\nabla V(f_t, z_t) = \frac{1}{|z_t|}\sum_{z\in z_t} \nabla V(f_t, z)$.

We  define a noise ``equivalent quantity''

\begin{equation}
	\xi_t =  \nabla V(f_t,z_t) - \nabla I_{S_n}(f_t),
\end{equation}
and it is clear that $\E\xi_t=0$.



We write Equation \ref{sgdK1} as

\begin{equation}\label{sgdK2}
f_{t+1}=f_t - \gamma_ t (\nabla I_{S_n} (f_t)+ \xi_t).
\end{equation}

\begin{figure*}[h!]\centering
\includegraphics[width=1.0\textwidth]{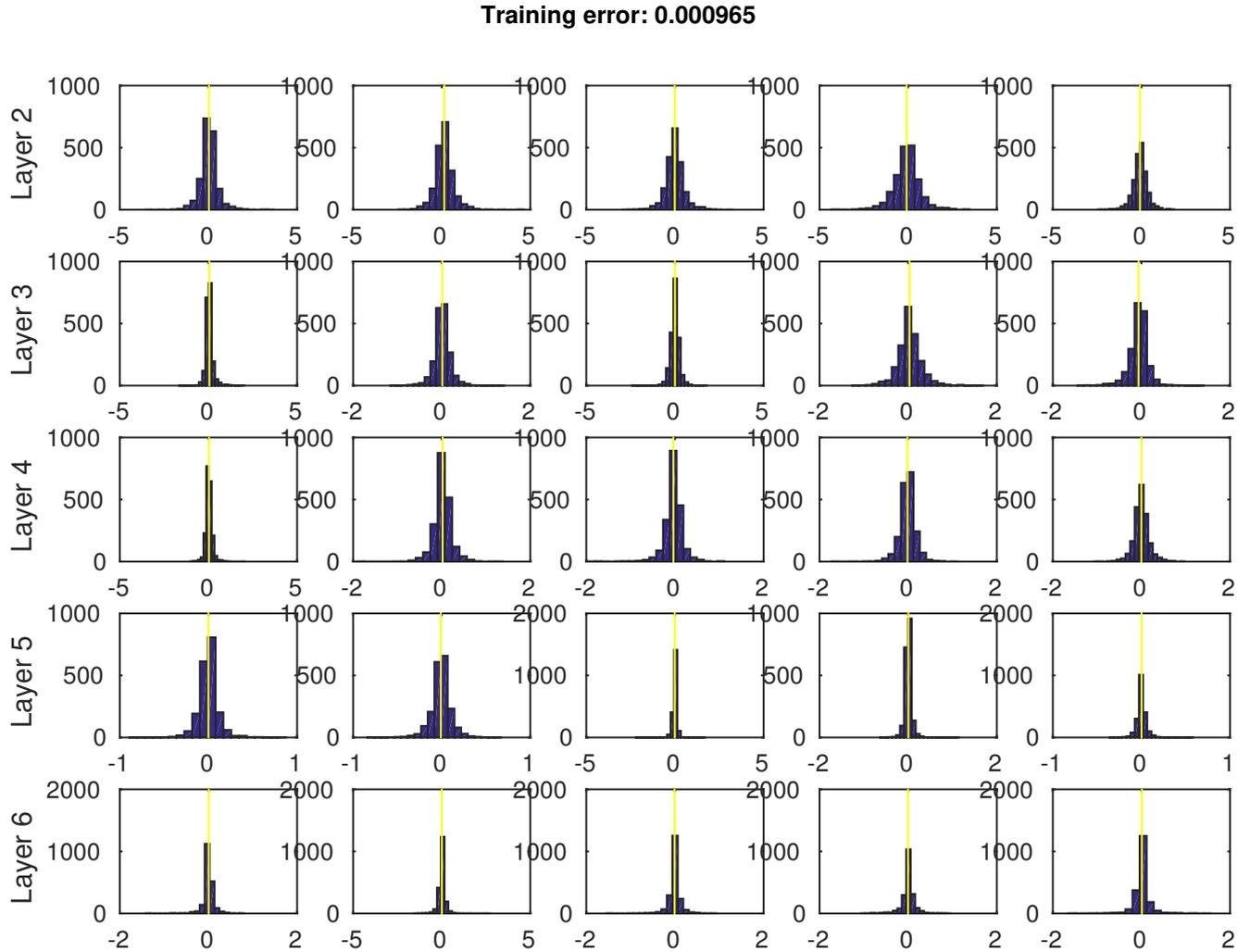}
\caption{Histograms of some of the components of $ \nabla V(f_t, z_i)$
  over $i$ for fixed $t$ in the asymptotic regime. Notice that the average corresponds to the
  gradient of the full loss and that is empirically very small. The
  histograms look approximatively Gaussian as expected (see text) for
  minibatches that are not too small or too large.}
\label{SGDgaussian}
\end{figure*}

With typical values used in minibatch (each minibatch corresponding to
$z_t$) training of deep nets, it turns out that the vector of gradient
updates $\nabla V(f_t, z_t)$ empirically shows components with
approximate Gaussian distributions (see Figure \ref{SGDgaussian}). This is expected because of the
Central Limit Theorem (each minibatch involves sum over many random
choices of datapoints).

Now we observe that \eqref{sgdK2} is a discretized Langevin diffusion,
albeit with a noise scaled as $\gamma_n$ rather than
$\sqrt{\gamma_n}$.  In fact, the continuous SGD dynamics corresponds
to a stochastic gradient equation using a potential function defined
by $U=I_{S_n}[f]= \frac{1}{n} \sum_{i=1}^n V(f, z_i)$ (see Proposition
3 and section 5 in \cite{bertsekas_gradient_2000}). If the noise were
the derivative of the Brownian motion, it is a Langevin equation --
that is a stochastic dynamical system -- with an associated
Fokker-Planck equation on the probability distributions of $f_t$. The
asympotic probability distribution is the Boltzman distribution that
is $\approx e^{\frac{-U}{\gamma K}}$.

For more details, see for instance section 5 of
\cite{Bertsekas:96}. Several proofs that adding a white noise term to
equation~\eqref{sgdK1} will make it converge to a global minimum are
available (see \cite{Gidas1985}).  Notice that the discrete version of
the Langevin dynamics is equivalent to a Metropolis-Hastings algorithm
for small learning rate (when the rejection step can be neglected).

\begin{figure*}[h!]\centering
\includegraphics[width=0.5\textwidth]{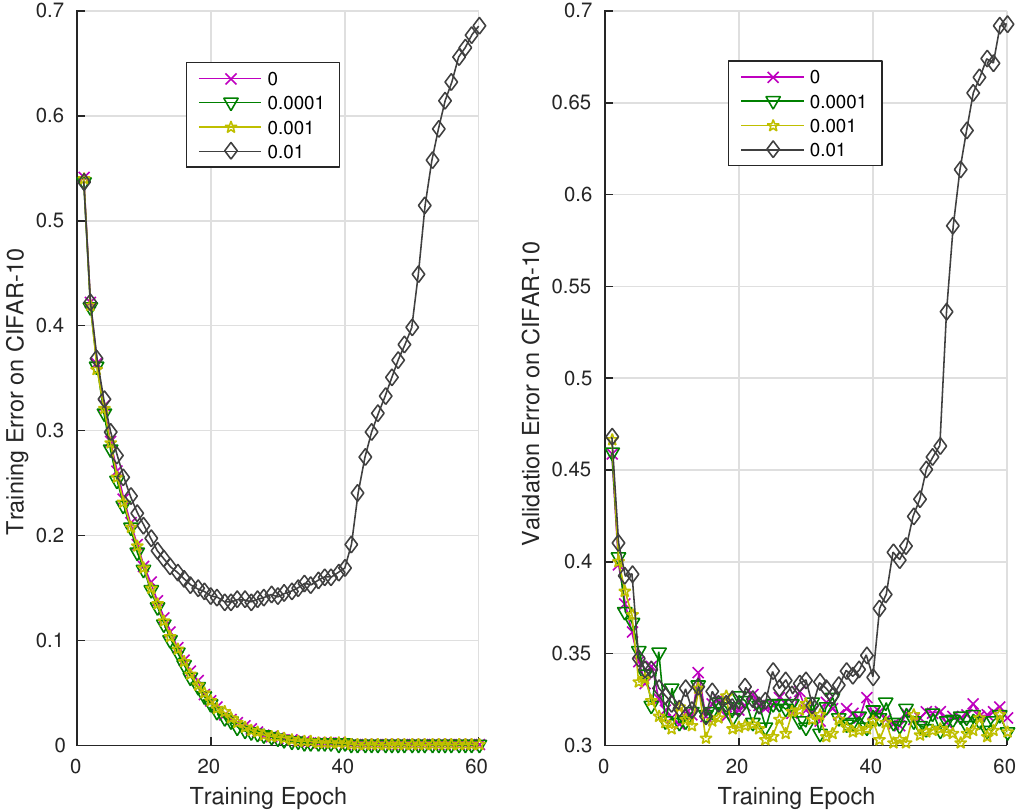}
\caption{Equation \ref{SGDL}  -- that is SGD with
  added Gaussian (with constant power) -- behaves in a similar way to
  standard SGD. Notice that SGDL has slightly better validation
  performance than SGD.}
\label{SGDLangevin}
\end{figure*}


\subsection{SGDL concentrates at large volume,
  ``flat'' minima}
\label{SGDLFlatRobust}
\begin{figure*}[h!]\centering
\includegraphics[width=1.0\textwidth]{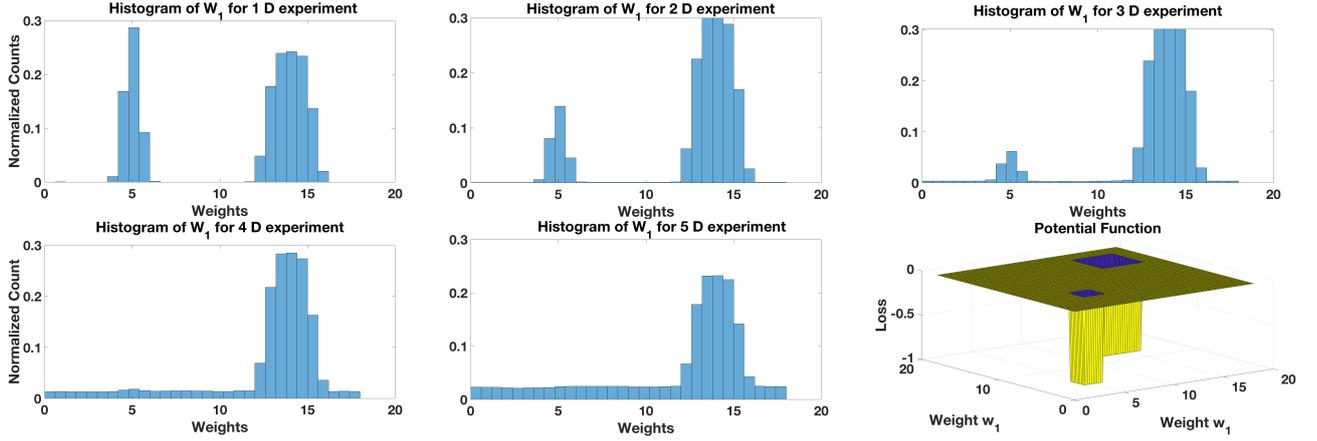}
\caption{The figure shows the histogram of a one-dimensional slice of
  the asymptotic distribution obtained by running Langevin Gradient
  Descent (GDL) on the potential surface on the right. The potential
  function has two minima: they have the same depth but one has a flat
  region which is a factor $2$ larger in each of the dimensions. The
  $1$D histogram for the first weight coordinate is shown here for
  dimensionality $1, 2, 3, 4$ and $5$D. The figures graphically show
  -- as expected from the asymptotic Boltzman distribution -- that
  noisy gradient descent selects with high probability minimizers with
  larger margin. As expected, higher dimensionality implies higher
  probability of selecting the flatter minimum.}
\label{large_margin_fig}
\end{figure*}

\begin{figure*}[h!]\centering
\includegraphics[width=0.9\textwidth]{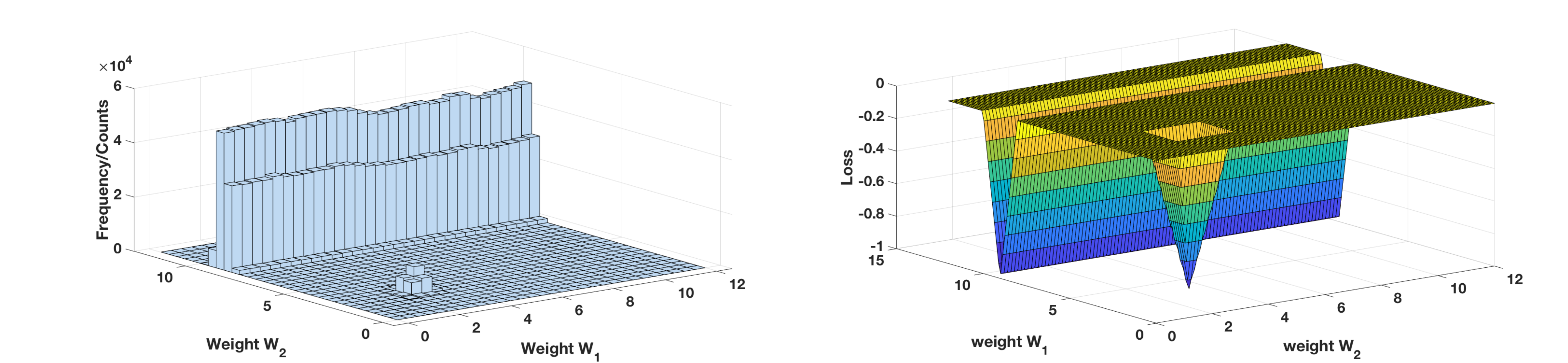}
\caption{Langevin Gradient Descent (GDL) on the $2$D potential function
  shown above leads to an asymptotic distribution with the
  histogram shown on the left. As expected from the form of the
  Boltzman distribution, the Langevin dynamics prefers degenerate
  minima to non-degenrate minima of the same depth. In high dimensions
we expect the asymptotic distribution to concentrate strongly around
the degenerate minima as confirmed on figure \ref{degenerate_fig_wedge_5D}.}
\label{degenerate_fig}
\end{figure*}

The argument about convergence of SGDL to large volume minima tha we
call ``flat'', is straighforward. The asymptotic distribution reached
by a Langevin equation (GDL) --as well as by SGDL -- is the Boltzman
distribution that is

\begin{equation}
p(f) = \frac{1}{Z}e^{-\frac{U}{T}},
\label{Bolzman}
\end{equation}

\noindent where $Z$ is a normalization constant, $U$ is the loss and
$T$ reflects the noise power. The equation implies, and Figure
\ref{degenerate_fig} shows, that SGD prefers degenerate minima
relative to non-degenerate ones of the same depth. In addition, among
two minimum basins of equal depth, the one with a larger volume, is
much more likely in high dimensions (Figure
\ref{large_margin_fig}). Taken together, these two facts suggest that
SGD selects degenerate minimizers and, among those, the ones
corresponding to larger isotropic flat regions of the loss. Suppose
the landscape of the empirical minima is well-behaved in the sense
that deeper minima have broader basin of attraction.Then it is
possible to prove that SDGL shows concentration -- {\it because of the
  high dimensionality} -- of its asymptotic distribution Equation
\ref{Bolzman} -- to minima that are the most robust to perturbations
of the weights.  Notice that these assumptions are satisfied in the
zero error case: among zero-minimizer, SGDL selects the ones that are
flatter, i.e. have the larger volume\footnote{Given a zero minimizer
  there is no other minimizer that has smaller volume AND is deeper.}.

In \cite{Musings2017} we will discuss how flat minima imply robust
optimization and maximization of margin.

\begin{figure*}[h!]\centering
\includegraphics[width=0.9\textwidth]{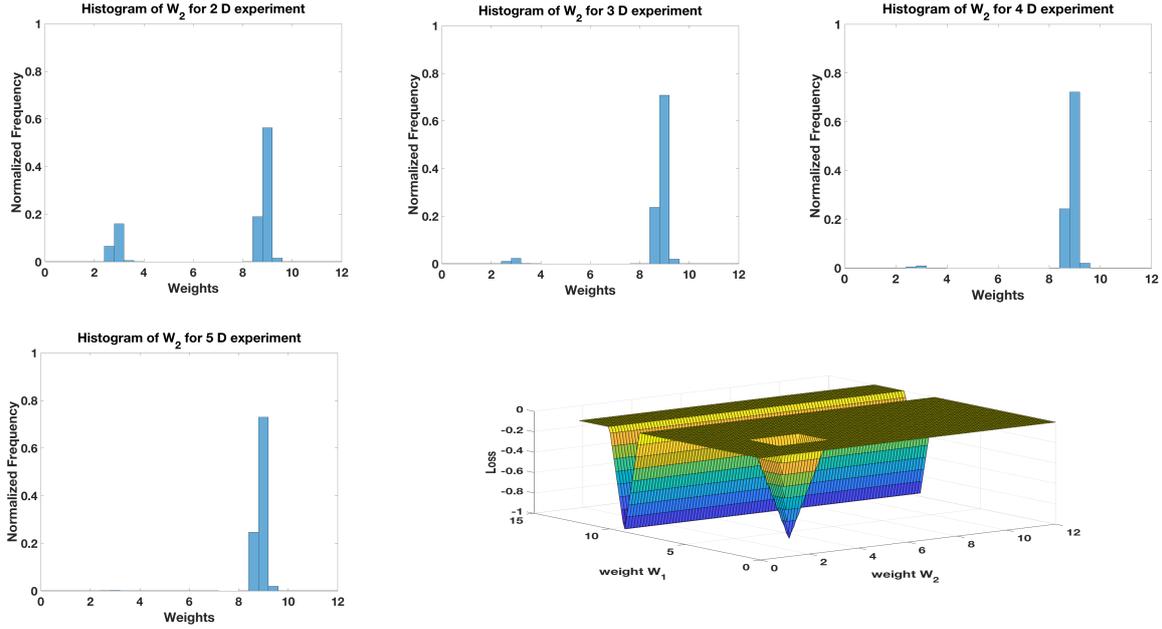}
\caption{The figure shows the histogram of a one-dimensional slice of
  the asymptotic distribution obtained by running Langevin Gradient
  Descent (GDL) on the potential surface on the right. As expected from the form of the
  Boltzman distribution, the Langevin dynamics prefers degenerate
  minima to non-degenrate minima of the same depth. Furthermore,
  as dimensions increase the distribution concentrates strongly around the degenerate minima.
  This can be appreciated from the figure because the histogram density at $W_1=2$ (the degenerate minimum)
  decreases in density rapidly as dimensions increases.}
\label{degenerate_fig_wedge_5D} 
\end{figure*}

\begin{figure*}[h!]\centering
\includegraphics[width=1.0\textwidth]{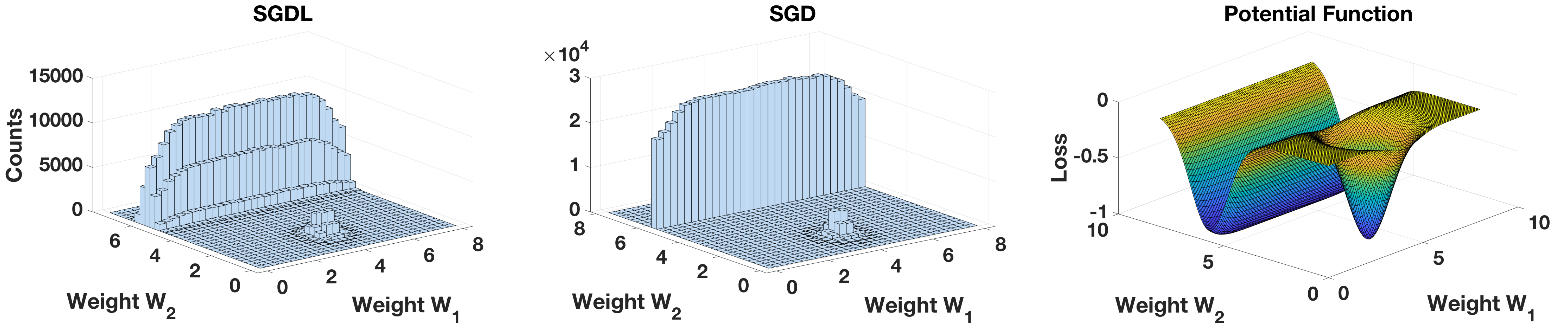}
\caption{  Stochastic Gradient Descent and Langevin Stochastic Gradient Descent (SGDL) on the $2$D potential function
  shown above leads to an asymptotic distribution with the histograms shown on the left. As expected from the form of the
  Boltzman distribution, both dynamics prefers degenerate minima to non-degenerate minima of the same depth.
}
\label{wedge_rbf_sgdl}
\end{figure*}

Here we observe that  {\it SGDL and SGD maximize volume
  and ``flatness'' of the loss in weight space}. Given a flat minimum,
one may ask where SGD will converge to.  For situations such as in
Figure \ref{degenerate_fig_wedge_5D} and for a minimum such as in
Figure \ref{wedge_rbf_sgdl}, Theory III suggests a locally minimum
norm solution. In particular,
the weight values found by SGD are expected to be mostly around their
average value over the flat region (at least in the case of square
loss). 

\section{Random Labels}
\label{RandomLabels}
For this case, Theory II predicts that it is in fact possible to
interpolate the data on the training set, that is to achieve zero
empirical error (because of overparametrization) and that this is in fact
easy -- because of the very high number of zeros of the polynomial
approximation of the network-- assuming that the
target function is in the space of functions realized by the
network. For $n$ going to infinity we expect that the empirical error
will converge to the expected (chance), as shown in the Figures
For finite $n $ when $n<W$, the fact that the empirical error (which
is zero) is so different from the expected seems puzzling, as observed by
\cite{zhang2016}, especially because the algorithm is capable of low
expected error with the same $n$ for natural labels.

\begin{figure}\centering
\begin{subfigure}[b]{.45\textwidth}
  \includegraphics[width=\textwidth]
  {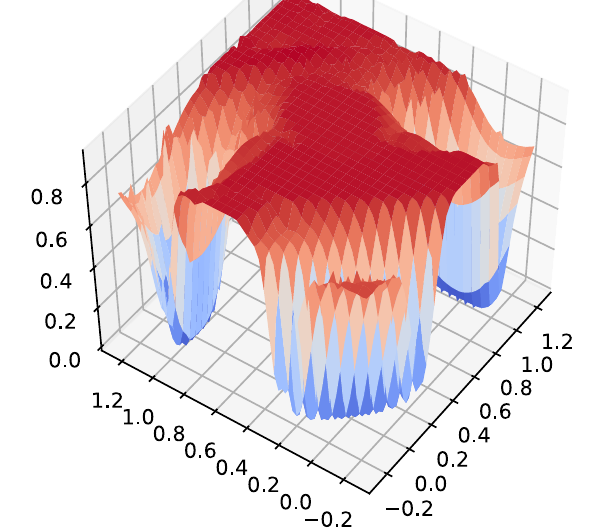}
  \caption{natural label}
\end{subfigure}
\hfill
\begin{subfigure}[b]{.45\textwidth}
  \includegraphics[width=\textwidth]
  {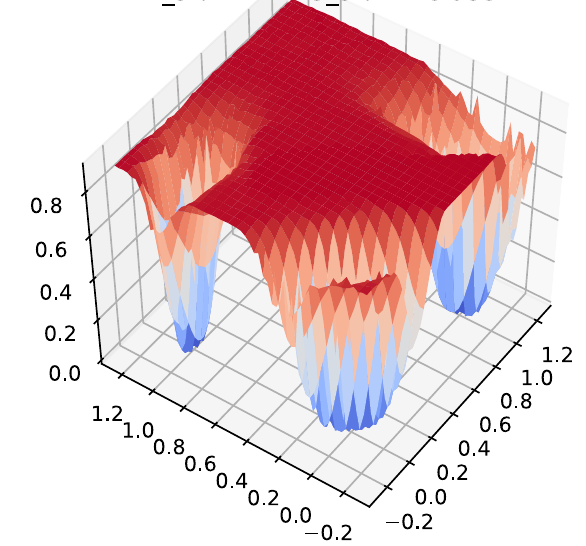}
  \caption{random label}
\end{subfigure}
\caption{Illustration of the landscape of the empirical loss on CIFAR-10.}
\label{fig:cifar10-3pt-interp}
\end{figure}

\begin{figure}\centering
\begin{subfigure}[b]{.45\textwidth}
  \includegraphics[width=\textwidth]
  {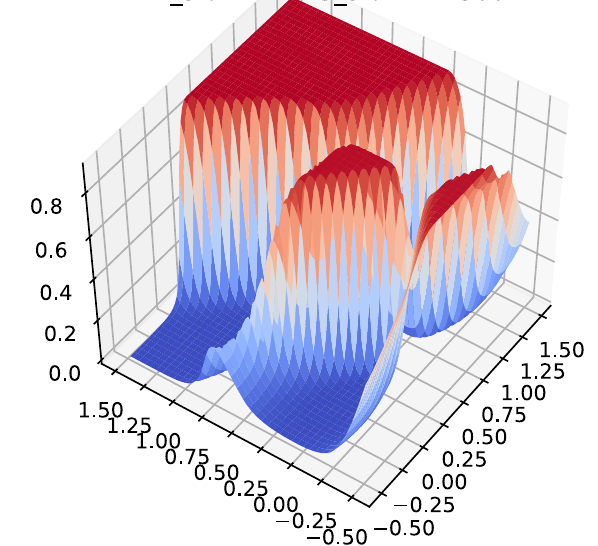}
  \caption{natural label}
\end{subfigure}
\hfill
\begin{subfigure}[b]{.45\textwidth}
  \includegraphics[width=\textwidth]
  {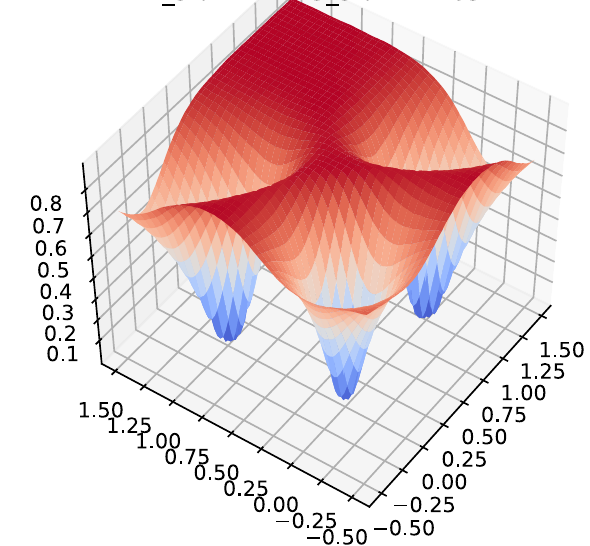}
  \caption{random label}
\end{subfigure}
\caption{Illustration of the landscape of the empirical loss on MNIST.}
\label{fig:mnist-3pt-interp}
\end{figure}

A larger margin is found for natural
labels than for random labels as shown in
Table~\ref{tab:isotropic-flatness-radius} and in
Figure~\ref{fig:cifar10-3pt-interp} and
Figure~\ref{fig:mnist-3pt-interp}.
Figure~\ref{fig:cifar10-3pt-interp} shows ``three-point
interpolation'' plots to illustrate the flatness of the landscape
around global minima of the empirical loss found by SGD, on CIFAR-10,
with natural labels and random labels, respectively. Specifically, let
$w_1, w_2, w_3$ be three minimizers for the empirical loss found by
SGD. For $\lambda=(\lambda_1,\lambda_2,\lambda_3)$ on the simplex
$\Delta_3$, let
\begin{equation}
  w_\lambda = \lambda_1 w_1 + \lambda_2 w_2 + \lambda_3 w_3
\end{equation}
We then evaluate the training accuracy for the model defined by each
interpolated weights $w_\lambda$ and make a surface plot by embedding
$\Delta_3$ in the 2D X-Y plane. As we can see, the natural label case
depict a larger flatness region around each of the three minima than
the random label case.  There is a direct
relation between the range of flatness and the norm $\lambda$ of the
perturbations. 

The same phenomenon could be observed more clearly on the MNIST
dataset, where the images of the same category are already quite
similar to each other in the pixel space, making it more difficult to
fit when random labels are used. Therefore, the difference in the
characteristics of the landscapes is amplified. As shown in
Figure~\ref{fig:mnist-3pt-interp}, big flat regions could be observed
in the natural label case, while the landscape for the random label
experiment resembles sharp wells.

It is  difficult to visualize the  flatness of the landscape when the weights
are typically in the scale of one million dimensions. To assess the isotropic flatness, we employ
the following procedure around a minimum found by SGD: choose a random isotropic direction
$\delta w$ with $\|\delta w\|=1$, perform a line search to find the ``flatness radius'' in that
direction:
\begin{equation}
  r(w,\delta w, \varepsilon) = \sup \{ r: |\hat{I}(w) - \hat{I}(w + r\delta w)| \leq \varepsilon \}
\end{equation}
The procedure is repeated $T$ times and the average radius is calculated. The overall procedure is
also repeated multiple times to test the average flatness at different
minima. The results are
shown in Table~\ref{tab:isotropic-flatness-radius}. For both CIFAR-10 and MNIST, we observe a
difference between the natural label and random label.

\begin{table}\centering
  \begin{tabular}{ccc}
  \hline
  & MNIST & CIFAR-10 \\\hline
  all params & $45.4 \pm 2.7$ & $17.0\pm 2.4$ \\
  all params (random label) & $6.9 \pm 1.0$ & $5.7\pm 1.0$ \\
  \hline
  top layer & $15.0\pm 1.7$ & $19.5\pm 4.0$ \\
  top layer (random label) & $3.0\pm 0.1$ & $12.1\pm 2.6$\\
  \hline
  \end{tabular}
  \caption{The ``flatness test'': at the minimizer, we move the weights around in a random
  direction, and measure the furthest distance until the objective function is increased by
  $\varepsilon$ (0.05), and then measure the average distance.}
  \label{tab:isotropic-flatness-radius}
\end{table}

\section{Discussion}

We expect the speed of convergence to be correlated
with good generalization because convergence will depend on the
relative size of the basins of attraction of the minima of the
empirical risk, which in turn depend on the ratio between the
effective dimensionality of the minima and the ambient
dimensionality. This paper, together with Theory II,
discusses  unusual properties of Stochastic
Gradient Descent used for training overparametrized deep convolutional
networks. SGDL and SGD select with high probability solutions with
zero or small empirical error (Theory II) -- because they are
flatter.

{\bf Acknowledgment}

This work was supported by the Center for Brains, Minds and Machines
(CBMM), funded by NSF STC award CCF – 1231216. We gratefully acknowledge the
support of NVIDIA Corporation with the donation of the DGX-1 used for
this research.


\bibliographystyle{ieeetr}
\small
\bibliography{Boolean}

\begin{thebibliography}{10}

\bibitem{Theory_I}
T.~Poggio, H.~Mhaskar, L.~Rosasco, B.~Miranda, and Q.~Liao, ``Why and when can
  deep - but not shallow - networks avoid the curse of dimensionality: a
  review,'' tech. rep., MIT Center for Brains, Minds and Machines, 2016.

\bibitem{Theory_II}
T.~Poggio and Q.~Liao, ``Theory ii: Landscape of the empirical risk in deep
  learning,'' {\em arXiv:1703.09833, CBMM Memo No. 066}, 2017.

\bibitem{keskar_large-batch_2016}
N.~S. Keskar, D.~Mudigere, J.~Nocedal, M.~Smelyanskiy, and P.~T.~P. Tang, ``On
  {Large}-{Batch} {Training} for {Deep} {Learning}: {Generalization} {Gap} and
  {Sharp} {Minima},'' {\em arXiv:1609.04836 [cs, math]}, Sept. 2016.
\newblock arXiv: 1609.04836.

\bibitem{dinh2017sharp}
L.~Dinh, R.~Pascanu, S.~Bengio, and Y.~Bengio, ``Sharp minima can generalize
  for deep nets,'' {\em arXiv preprint arXiv:1703.04933}, 2017.

\bibitem{GelMit91}
S.~Gelfand and S.~Mitter, ``Recursive stochastic algorithms for global
  optimization in {$R^d$},'' {\em Siam J. Control and Optimization}, vol.~29,
  pp.~999--1018, September 1991.

\bibitem{bottou-98x}
L.~Bottou, ``Online algorithms and stochastic approximations,'' in {\em Online
  Learning and Neural Networks} (D.~Saad, ed.), Cambridge, UK: Cambridge
  University Press, 1998.
\newblock revised, oct 2012.

\bibitem{bertsekas_gradient_2000}
D.~Bertsekas and J.~Tsitsiklis, ``Gradient {Convergence} in {Gradient} methods
  with {Errors},'' {\em SIAM J. Optim.}, vol.~10, pp.~627--642, Jan. 2000.

\bibitem{Bertsekas:96}
D.~P. Bertsekas and J.~N. Tsitsiklis, {\em Neuro-dynamic Programming}.
\newblock Athena Scientific, Belmont, MA, 1996.

\bibitem{Gidas1985}
B.~Gidas, ``{Blobal optimization via the Langevin equation},'' {\em Proceedings
  of the 24th IEEE Conference on Decision and Control}, pp.~774--778, 1985.

\bibitem{Musings2017}
C.~Zhang, Q.~Liao, A.~Rakhlin, K.~Sridharan, B.~Miranda, N.Golowich, and
  T.~Poggio, ``Musings on deep learning: Optimization properties of sgd,'' {\em
  CBMM Memo No. 067}, 2017.

\bibitem{zhang2016}
C.~Zhang, S.~Bengio, M.~Hardt, B.~Recht, and O.~Vinyals, ``Understanding deep
  learning requires rethinking generalization,'' in {\em International
  Conference on Learning Representations (ICLR)}, 2017.

\end{thebibliography}
\normalsize

\end{document}